# Intention-aware Long Horizon Trajectory Prediction of Surrounding Vehicles using Dual LSTM Networks

Long Xin[1,2], Pin Wang[1], Ching-Yao Chan[1], Jianyu Chen[3], Shengbo Eben Li[2*], Bo Cheng[2],

*Abstract*—As autonomous vehicles (AVs) need to interact with other road users, it is of importance to comprehensively understand the dynamic traffic environment, especially the future possible trajectories of surrounding vehicles. This paper presents an algorithm for long-horizon trajectory prediction of surrounding vehicles using a dual long short term memory (LSTM) network, which is capable of effectively improving prediction accuracy in strongly interactive driving environments. In contrast to traditional approaches which require trajectory matching and manual feature selection, this method can automatically learn high-level spatial-temporal features of driver behaviors from naturalistic driving data through sequence learning. By employing two blocks of LSTMs, the proposed method feeds the sequential trajectory to the first LSTM for driver intention recognition as an intermediate indicator, which is immediately followed by a second LSTM for future trajectory prediction. Test results from real-world highway driving data show that the proposed method can, in comparison to state-of-art methods, output more accurate and reasonable estimate of different future trajectories over 5s time horizon with root mean square error (RMSE) for longitudinal and lateral prediction less than 5.77m and 0.49m, respectively.

## I. INTRODUCTION

Autonomous driving are believed to have the potential for improvement of road safety, energy efficiency, traffic congestion and drivers' relief from driving burden. To guarantee driving safety in dynamic traffic, an autonomous vehicle (AV) should be able to anticipate the traffic environment in the future and respond to these changes appropriately. However, the motion of traffic participants, especially those surrounding the AV, is often difficult to predict since it is affected by various factors such as road user interaction, randomness of driver behavior and road geometry constraints. In addition, the observation from surrounding vehicles (SVs) are usually imperfect and highly noisy due to environment complexity and performance of commonly available low-cost sensors. Thus, the assumption of full knowledge of the surrounding vehicles' states, i.e., heading angle, yaw rate, etc., is not feasible without support from other sources, such as the cooperative concept of V2X through wireless communication among vehicles. Thus, it is critically important to implement efficient and effective trajectory prediction of surrounding vehicles to cope with challenges mentioned above, so as to facilitate the decision-making and path planning system for higher level of autonomous driving.

Implemented with object tracking techniques, vehicle motion model such as kinematic or dynamic models has been used for trajectory prediction [1][2]. Since the observation is often limited, Kalman filter has been widely used for prediction by taking the uncertainty in vehicle model into consideration [1]. In order to account for more influencing factors and improve the prediction accuracy further, Bayesian filtering techniques such as the context-dependent interactive multiple model filter [3] and Monte-Carlo method [2] have been proposed. These methods, while taking into account the physical limitations of a vehicle, are normally effective for short-term trajectory prediction, i.e. one or two seconds in the future, and are not accurate enough for long-term prediction, which directly affects the accuracy of decision making and path planning.

An popular alternative to the challenges mentioned above is to take advantage of the prototype trajectory so that prediction can be performed by comparing the current trajectory with the learned motion patterns and using the prototype trajectory as a base model for future motion [4]. These prototype vehicle trajectories can be learned through Gaussian process [5][6] and Gaussian mixture model [7]. The downside of Gaussian model is the expensive online computation load of calculating the probability similarity of the current trajectory with the prototype one. Besides, it is time dependent, i.e. trajectories falling into the waiting intervals when a vehicle stops have to be manually dropped out [8][9]. When motion patterns are represented by a finite set of prototype trajectories, the similarity of a partial trajectory to a motion pattern is measured by metrics such as the Longest Common Subsequence (LCS) [10], the average Euclidian distance between points of the trajectories [11], the modified Hausdorff [12], etc. A disadvantage of such method is that using a finite set of trajectories would take a very large number of prototypes to model various patterns of the real-world driving trajectories. Another difficulty is the adaption to different road geometry, as the learned prototype trajectory models can only be applied in a similar road layout.

Deep neural networks (DNN) have recently gained increasingly popularity as universal function approximators, capable of learning hierarchical and semantic features from complicated inputs [13][14] and outperformed traditional methods in fields from image classification [15] to natural language processing [16]. Due to the unique neural network structure, recurrent neural network (RNN) is widely used to learn the temporal structure of time-series data.

[1] L. Xin, P. Wang, and C. Chan are with California PATH, University of California, Berkeley, CA 94804, USA. (email: {long_xin, pin_wang, cychan}@berkeley.edu)
[2] L. Xin, S.E. Li, and B. Cheng are with State Key Lab of Automotive Safety and Energy, Department of Automotive Engineering, Tsinghua University, Beijing 100084, China. (email: xinlong.thu@gmail.com, {lishbo, chengbo} @tsinghua.edu.cn)
[3] J. Chen is with Department of Mechanical Engineering, University of California, Berkeley, CA 94720, USA. (email: jianyuchen@berkeley.edu)
* All correspondence should be sent to S.E. Li.

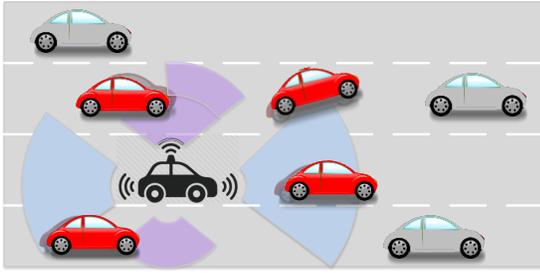

Fig. 1. Limited observation of SVs with occlusion and noise

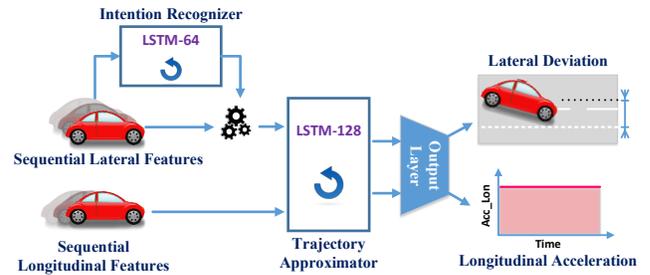

Fig. 2. Proposed system architecture

In particular, the long short term memory (LSTM) architecture has been successfully applied to analyze the temporal structure underlying in text, speech, and financial data [17], as well as lateral position prediction [18] and driver intent prediction [19] in the field of intelligent driving. Reference [20] directly applied LSTM to predict future trajectory over occupancy grid. Although the occupancy grid helps to deal with the prediction uncertainty, the method sacrifices prediction accuracy in terms of position as the grid is 10m long. Reference [21] proposed an LSTM network for highway trajectory prediction that leveraged the information of surrounding vehicles around the target vehicle. The main downside, however, is the infeasibility of observing these information for a target vehicle which is already a surrounding vehicle of the ego vehicle due to sensing occlusion. Again, similar to prototype trajectories, these studies are not capable of dealing with different road layouts in a universal way as it is difficult to tell a lane change behavior from driving along a curved road, thus misleading the prediction results.

In contrast to the existing methods, this paper proposes a more pragmatic and geography-adaptive method which is first to estimate the maneuver intention of the driver and then to predict the successive kinematic states in future horizon to correspond with the possible execution of the identified maneuver. In our work, we employ two LSTMs to recognize high-level driver intention as well as to understand low-level complex dynamics of vehicle motion. The proposed trajectory prediction system inputs the sequentially transformed coordinates of the surrounding vehicles obtained from the sensor measurements to the LSTMs and produces the vehicle's future locations in the next 5 seconds. The method is developed based on naturalistic driving data and shows better prediction accuracy over some existing methods.

The rest of this paper is organized as follows. In Section II, the problem is formally defined. In section III, the proposed method and its related techniques are introduced. The dataset used and the training method employed are described in Section IV. In Section V, the results are provided and the paper is concluded in Section VI.

## II. PROBLEM STATEMENT

To achieve better accuracy in decision making and path planning modules of an autonomous vehicle, it is beneficial to address the problem of predicting future trajectories of surrounding vehicles on a highway using historic but limited observation data from on-board sensors. Different from the work in [20] where vehicle trajectories are represented in the form of occupancy grid and turned the prediction task into a classification problem, we consider it as a regression one whose outputs are supposed to be as accurate as actual values.

More formally, our goal is to train a predictor for future trajectory of sequential outputs $Y = \{y_t\}_{t \in T_{post}, y \in \mathcal{O}}$ based on a set of observable feature inputs $X = \{x_t^k\}_{t \in T_{prev}, x^k \in \mathcal{F}}$, where $T_{prev} = \{-t_{pres}, \dots, 0\}$ and $T_{post} = \{1, \dots, t_{post}, \}$ are respectively the time intervals of the historical input and future predictions, $\mathcal{O}$ is the target outputs and $\mathcal{F}$ is the feature set acquired and calculated simultaneously. For $x^k \in \mathcal{F}$ and $t \in T_{post}$, we denote $x_t^k$ by the value of feature $x^k$ observed $t$ time steps earlier. Similarly, we denote $y_t$ by the value of output $t$ time steps in the future.

In this paper, future trajectories of surrounding vehicles to be predicted only depend on information observed from the ego vehicle which is usually partially observable due to sensor limitations and object occlusions. In contrast to [21] which used information of vehicles around a target surrounding vehicle to the ego vehicle, our assumption on the data availability is more pragmatic for real driving scenarios. In order to overcome the challenges of information limitation, data imbalance between lane keeping and other driving maneuvers, and model generalization, a novel framework based on intention-aware LSTM network is proposed and presented in Section III.

## III. PROPOSED METHOD

### A. System Architecuture

In this paper, the proposed deep RNN network is presented in Fig. 2. It mainly consists of two LSTM networks, one with a layer of 64 cells and the other with a layer of 128 cells, followed by a dense (fully connected) output layer containing as many neurons as the number of outputs. In this architecture, the role of the first LSTM layer is to recognize drivers' intentions, e.g. lane keeping and lane change, and transform the lateral features while the second LSTM layer is to extract meaningful representation of the sequential inputs and thus produce future trajectories with help of the dense layer and the embedded semantic understanding of driving intentions.

Compared to the existing studies [20][21], this architecture enjoys several advantages. One is that the semantic understanding of driving intention can be used as instructive information for inferring surrounding vehicles' motion. Another is that upper and lower boundaries for lateral trajectory predictions can be generated based on prior knowledge of driving code on structured road, i.e. driving along lane centers, changing one lane at a time, etc. Moreover, the delay of trajectory prediction when a vehicle is changing lanes can be minimized with the help of timely intention recognition. The last but not least advantage is that both lane keeping and lane change data can be treated under a unified framework during the learning process and the negative effect of the imbalance of data can be reduced (see Section III-B). These advantages will help improve the accuracy of long-horizon prediction outputs under the constraint of limited observation data.

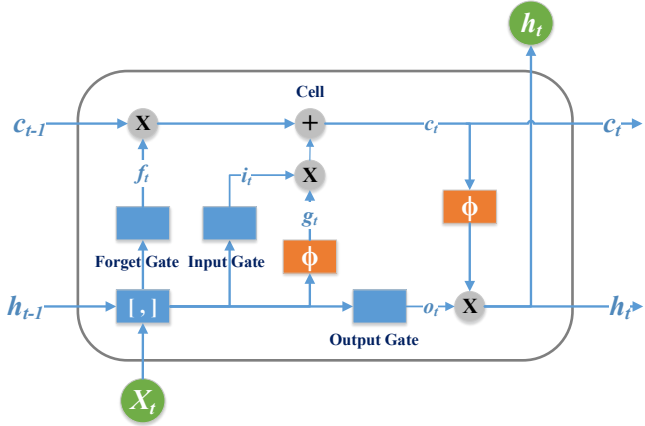

Fig. 3. Internal sturcture of an LSTM cell

### B. Input Features

From the view point of the ego vehicle, only the features that can be feasibly measured using on-board sensors, such as LiDAR and radar, are used as input. Similarly, all the input features are measured and represented in the local coordinate system parallel to the global one as the AV moves. For each surrounding vehicle, its relative lateral position is defined as

$$x_{rel} = x - x_n \quad (1)$$

where $x$ is local lateral position and $x_n$ is local lateral position of the nearest lane marking. This feature along with the first and the second derivative, $\dot{x}_{rel}$ and $\ddot{x}_{rel}$, are helpful for driving intention recognition especially when analyzing them in a time series.

Similarly, its lateral deviation from the center line of the target lane is defined as

$$x_{dev} = x - x_{targ} \quad (2)$$

where $x_{targ}$ is the local lateral position of the center line of the target lane to which the surrounding vehicle is heading.

The target lane is defined according to the driving intention recognized from the first LSTM network. If the recognized intention is driving along the lane, then the target lane is the current lane. If the intention is to change lane, then the target lane is the neighboring one that it intends to change to. The benefits of this setting lie in the ability of adaption to the road geometry and scalability of the range of feature values bounded to the activation functions.

In addition to the 4 features introduced above, lateral features $\dot{x}_{dev}$ and $\ddot{x}_{dev}$ and longitudinal features $v_y$ and $a_y$ are also adopted as input features. These features are selected not only based on the sensor observation but also the decision factors a human driver is likely to rely on.

### C. LSTM Network

RNNs are distinguished from traditional feed-forward networks by its internal states and cycles, which are capable of analyzing sequential information and learning temporal features. The LSTM, a particular implementation of RNN, was developed to avoid vanishing gradients of the loss function over time [17].

The LSTM has a memory state called 'cell' which stores the interpretation of past input data. The cell is updated based on the current input and the previous cell state. Between the input and the cell are different gates, a unique control mechanism that enables the LSTM to learn when to forget past state and update the state when given new input. Let $c_t$ be memory cell state at time step $t$ and $h_t$ be the output hidden state, then $c_t$ and $h_t$ are updated by the following equations

$$i_t = \sigma(w_{x^i}x_t + w_{h^i}h_{t-1} + b_i) \quad (3)$$
$$f_t = \sigma(w_{x^f}x_t + w_{h^f}h_{t-1} + b_f) \quad (4)$$
$$o_t = \sigma(w_{x^o}x_t + w_{h^o}h_{t-1} + b_o) \quad (5)$$
$$g_t = tanh(w_{x^c}x_t + w_{h^c}h_{t-1} + b_c) \quad (6)$$
$$c_t = f_t \odot c_{t-1} + i_t \odot g_t \quad (7)$$
$$h_t = o_t \odot \tanh(c_t) \quad (8)$$

where $\sigma(x) = \frac{1}{1+e^{-x}}$ is the sigmoid function; $i_t$, $f_t$, $o_t$ and $g_t$ are input gate vector, forget gate vector, output gate vector and state update vector, respectively; $w_{x^i}$, $w_{x^f}$, $w_{x^o}$, $w_{x^c}$, $w_{h^i}$, $w_{h^f}$, $w_{h^o}$, $w_{h^c}$ are the weights for linear combination; $b_i$, $b_f$, $b_o$ and $b_c$ are the relative bias; $\odot$ is element-wise production.

Fig. 3 shows the internal structure of an LSTM cell. The input gate $i_t$ and the output gate $o_t$ can control the data flow from the input and to the output, respectively. The forget gate $f_t$ can decide whether to forget the information stored in the memory cell. The gating mechanism is learned from data so that problems in training caused by the input decaying or increasing exponentially over time in general RNNs can be overcome. Note that the LSTM is considered as a deep neural network when unfolded along time. Addition LSTM layers can be added to extract even higher level of features and make the network deeper. In this paper, however, similar results were found using different layers of LSTMs, therefore only one layer of LSTM is finally used to save storage and computing resource. Note that an additional fully-connected dense output layer to the hidden state is added to generate outputs in different forms relevant to the given task, i.e., intention recognition and trajectory prediction.

### D. Prediction Outpus

The region of interest spans roughly 500 meters longitudinally, the values of longitudinal positions can be quite large. In order to predict future trajectories of surrounding vehicles, future longitudinal accelerations over the prediction horizon $\{\hat{a}_y(t)\}_{t \in T_{post}}$ are output instead.

Similar to [24], the future longitudinal trajectory is then calculated accordingly:

$$\hat{v}_y(t) = \hat{v}_y(t-1) + \hat{a}_y(t) * \delta t \quad (9)$$

$$\hat{y}(t) = \hat{y}(t-1) + \hat{v}_y(t) * \delta t \quad (10)$$

where $\delta t$ is the time step.

Since the lateral position is bounded, the lateral deviation over the prediction horizon $\{\hat{x}_{dev}(t)\}_{t \in T_{post}}$ are directly used for the output. Then the lateral position can be calculate as

$$\hat{x}(t) = \hat{x}_{dev}(t) + \hat{x}_{targ} \quad (11)$$

where $\hat{x}_{targ}$ is the local lateral position of the centerline of the inferred target lane based on the intention recognition output from the first LSTM layer.

## IV. DATA AND TRAINNING

### A. Data

The dataset used in this paper is from the Next Generation Simulation (NGSIM) [22]. Collected and published by the US Federal Highway Administration in 2005, the NGSIM is one of the largest open datasets of naturalistic driving and has been widely studied in the literature, e.g., [21][23][24].

More specifically, the area of interest is the I-80 freeway in Emeryville, California, of which the covered segment is approximately 500m in length and 6 lanes (3.66m or 12ft each) in width (see Fig. 2). The 45-minute trajectory data were collected from 4:00pm to 4:15pm and from 5:00pm to 5:30pm, reflecting different traffic characteristics during transitional and congested traffic period, respectively.

The dataset contains more than 5000 trajectories of individual vehicles, with a sampling rate at 10 Hz. Each sample in one trajectory includes the information such as instantaneous speed, acceleration, longitudinal and lateral positions, vehicle length, and vehicle type. The local coordinates is set at the down-left point of the study area, where $x$ is the lateral position of the vehicle relative to the leftmost edge of the road, and $y$ its longitudinal position to the entry edge. Furthermore, a total of 914 successful lane changes were identified automatically per SAE J2944 [25], while the lane change to the left totaled 694.

### B. Training

The proposed network is trained using a window of 50 time steps, representing 5s past observations. Such window is updated every second at 10 Hz. To increase training efficiency and avoid back propagation related issues, input data are grouped by batches of 100 and shuffled within batches.

For the first LSTM layer, it can be interpreted as a multi-class classification problem. Driving intentions, namely lane keeping, left lane change and right lane change, are encoded into a 1x3 vector with one hot technique when training. A softmax layer is added to the LSTM output with the linear transformation of hidden state $h_t$ followed by the softmax function

$$P(y = j | h_t) = \frac{e^{h_t^T w_j}}{\sum_{i=1}^{k} e^{h_t^T w_i}} \quad for \ j = 1, \dots, k \quad (12)$$

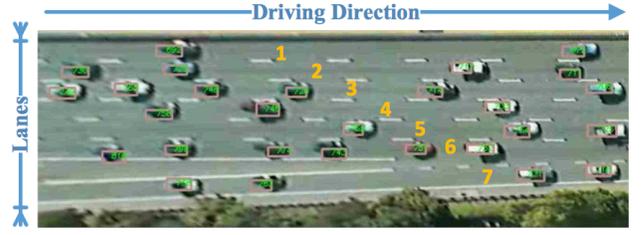

Fig. 4. Birdview of naturalistic traffic recorded on I-80 freeway

where $y$ is the driving intention prediction; $w_i$ are the weights related to each intention category; $k = 3$ is the number of intention categories.

For the second LSTM layer, defined features are directly taken as inputs and the loss function is set as the quadratic form of difference between the network predictions and the actual values. $L_2$ loss is calculated and weights are updated through back propagation through time (BPTT).

## V. RESULTS

This paper explored the long-time horizon (up to 5s) prediction of future vehicle trajectories for different driving intentions based on LSTMs. Samples are from the NGSIM I-80 dataset (70% for training, 30% for validation). The training is performed on GPU using the TensorFlow for 5 epochs. The initial learning rate is set to 1.0 and gradually decreases until the validation error stops to improve.

For the test, the input features of each trajectory are directly fed to the network, without additional process, e.g., grouping the data by time windows and priori selection of specific trajectory segments. The proposed network takes features extracted from past trajectory as input, recognize the driver intention on a probability basis and simultaneously predict future trajectory as output. Motion predictions of three different vehicles (identified as 831, 967 and 928) are presented in Fig. 5, with time difference of 2s. Their trajectories are represent by solid lines, dashed lines and dotted lines, respectively, with predicted trajectories in red and real ones in yellow. The proposed method analyzes and adapt itself to individual driving characteristics as the predicted trajectories differ from each other. Compared through Fig. 5 (a) to (c), a lane change action of vehicle 831 is recognized in time and in advance on a probability basis in the background system. With the help of intention recognition, the network is able to make human-like judgment of the motion pattern of surrounding vehicles, thus providing a good guidance for the low-level trajectory predictor.

For each individual trajectory, the Root Mean Squared Error (RMSE) between the predicted and the actual value is then computed to assess the learning performance of the proposed network as well as the generalization capability over different drivers. Results using other LSTM-based methods are introduced for comparison purposes. One is proposed by [20] which predicts future trajectory over occupancy grid. The other one is proposed by [21] which leverages information of vehicles around target vehicles.

Our proposed model provides the best overall results for longitudinal position prediction and similar results to the best prediction for lateral position. Note that [21] takes much more

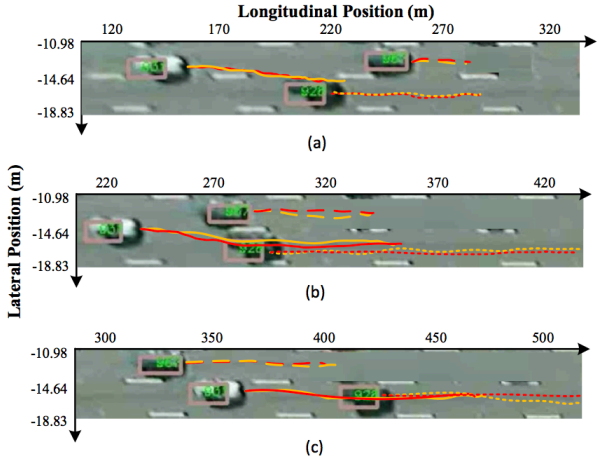

Fig. 5. Examples of typical trajectory prediction over 2s (Trajectories of vehicle 831, 967 and 928 are represent by solid lines, dashed lines and dotted lines, respectively, with predicted trajectories in red and real ones in yellow.)

TABLE I. RMSE FOR TRAJECTORY PREDICTIONS

| Model | Prediction Horizon | | | | |
|---|---|---|---|---|---|
| | 1s | 2s | 3s | 4s | 5s |
| Proposed Network | 0.47 | 1.39 | 2.57 | 4.04 | 5.77 |
| Altché et al. [21]* | 0.71 | 1.98 | 3.75 | 5.96 | 9.00 |
| Kim et al. [20]** | 3.05 | 6.70 | - | - | - |

(a) Longitudinal position error (m)

| Model | Prediction Horizon | | | | |
|---|---|---|---|---|---|
| | 1s | 2s | 3s | 4s | 5s |
| Proposed Network | 0.15 | 0.26 | 0.38 | 0.45 | 0.49 |
| Altché et al. [21]* | 0.11 | 0.25 | 0.33 | 0.40 | 0.47 |
| Kim et al. [20]** | 0.56 | 1.24 | - | - | - |

(b) Lateral position error (m)

vehicle information. Therefore, compared to other models, the proposed model extracts semantic information of driving intention which helps considerably to narrow down prediction boundaries even though the observation is limited. As for the longitudinal position prediction, predicting acceleration seems better than predicting velocity and position directly, probably due to the reason that driving patterns rely much on acceleration characteristics. Note that the RMSE calculation of [21] and [20] has been modified accordingly to position errors in SI units (denoted by a * and a ** in Table I).

Besides the RMSE metric, discrete events such as lane changes have been studied to evaluate the quality of prediction from another perspective. Since straight-line driving trajectories overwhelmingly exist in this dataset, lane change events is highly outnumbered and prediction errors of such events are not properly accounted for using RMSE, especially for longer prediction horizons. However, lane change predictions are so important that they have even greater impact on the safety issues of the ego vehicles and should be taken into careful consideration when making decisions and planning paths. Unlike the limitation faced by [21] which can cause huge prediction delay up to 8s or 9s, the proposed model can recognize driver intention seconds before the vehicle crosses the lane marking, which explicitly seems as a lane change behavior, thus reduce prediction delay to the maximum extent. Therefore, the relations between the probability of lane change events are correctly recognized and the time to the lane change point where the vehicles actually cross the lane marking are computed for left lane change events (LLC) and right lane change events (RLC), respectively. As shown in Fig. 6, all the lane change events can be recognized in advance which enables the proposed model to overcome the imbalance of the dataset in terms of driving maneuvers.

## VI. CONCLUSION

In this paper, a deep neural network architecture based on intention-aware LSTMs is proposed to predict long term (5s) vehicle trajectories on highway. Training and test using the widely studied NGSIM dataset shows better prediction accuracy and prediction timeliness, with outcomes indicated by smaller RMSE for both longitudinal and lateral position prediction over different time horizons and minimized prediction delay of lane change events. Compared to the state-of-art studies, the proposed model has the following advantages: (1) small prediction bounds for lateral position to guarantee prediction accuracy and feasibility; (2) not subject to internal imbalance of dataset; (3) adaptability to various road geometry due to the special representation of lateral deviation from the center of the target lane; and (4) providing guidance to high level decision making systems with semantic understanding of surrounding vehicles' motion.

The work done in this paper provides a promising way to handle motion prediction of surrounding vehicles for autonomous driving systems. The preliminary research opens various perspectives for future research: (1) to generalize the proposed model in different scenarios, e.g. intersections and unstructured roads; (2) to consider surrounding vehicles as a whole which reflects interactive relation between the ego vehicle and other vehicles; (3) to address the motion prediction as a stochastic problem which requires distributions and confidence intervals instead of just a single value.

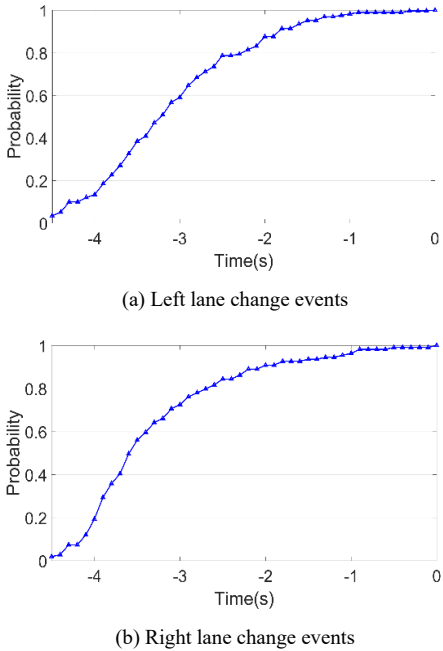

Fig. 6. Distribution of lane change intention recognition probability along time axis for all lane change events